\documentclass{article}

\usepackage[preprint]{neurips_2019}

\usepackage[utf8]{inputenc} 
\usepackage[T1]{fontenc}    
\usepackage{hyperref}       
\usepackage{url}            
\usepackage{booktabs}       
\usepackage{amsfonts}       
\usepackage{nicefrac}       
\usepackage{microtype}      

\usepackage{graphicx}
\usepackage{amsopn}
\usepackage{amsmath}
\usepackage{subcaption}
\usepackage{dsfont}
\usepackage[linesnumbered,ruled]{algorithm2e}

\usepackage{listings}
\usepackage{color}
\usepackage{placeins}

\usepackage{xcolor}
 
\definecolor{codegreen}{rgb}{0,0.6,0}
\definecolor{codegray}{rgb}{0.5,0.5,0.5}
\definecolor{codepurple}{rgb}{0.58,0,0.82}
\definecolor{backcolour}{rgb}{0.95,0.95,0.92}

\lstdefinestyle{mystyle}{
    backgroundcolor=\color{backcolour},
    commentstyle=\color{codegreen},
    keywordstyle=\color{magenta},
    numberstyle=\tiny\color{codegray},
    stringstyle=\color{codepurple},
    basicstyle=\tiny,
    breakatwhitespace=false,
    breaklines=true,
    captionpos=b,
    keepspaces=true,
    numbersep=5pt,
    showspaces=false,
    showstringspaces=false,
    showtabs=false,
    tabsize=1,
    xleftmargin=0pt,
    xrightmargin=0pt,
}
 
\title{A Projectional Ansatz to Reconstruction}

\author{%
  S\"oren Dittmer \\
  Center for Industrial Mathematics (ZeTeM) \\
  University of Bremen, Germany \\
  \texttt{soeren.dittmer@gmail.com} \\
  \And
  Peter Maass \\
  Center for Industrial Mathematics (ZeTeM) \\
  University of Bremen, Germany \\
  \texttt{pmaass@uni-bremen.de} \\
}

\begin{document}
\maketitle

\begin{abstract}
Recently the field of inverse problems has seen a growing usage of mathematically only partially understood learned and non-learned priors. Based on first principles, we develop a projectional approach to inverse problems that addresses the incorporation of these priors, while still guaranteeing data consistency. We implement this projectional method (PM) on the one hand via very general Plug-and-Play priors and on the other hand, via an end-to-end training approach. To this end, we introduce a novel alternating neural architecture, allowing for the incorporation of highly customized priors from data in a principled manner. We also show how the recent success of Regularization by Denoising (RED) can, at least to some extent, be explained as an approximation of the PM. Furthermore, we demonstrate how the idea can be applied to stop the degradation of Deep Image Prior (DIP) reconstructions over time.
\end{abstract}

\section{Introduction}
Recently the field of inverse problems has seen a growing usage of mathematically only partially understood learned (e.g., \citet{lunz2018adversarial, adler2018learned, hauptmann2018model, yang2018dagan, bora2017compressed}) and non-learned (e.g., \citet{venkatakrishnan2013plug, romano2017little, ulyanov2018deep, van2018compressed, mataev2019deepred, dittmer2018regularization}) priors.

A key challenge of many of these approaches, especially the learned ones, lies in the fact that it is often hard or even impossible to guarantee data consistency for them, i.e., that the reconstruction is consistent with the measurement, a notable exception to this is\ \cite{schwab2018deep} in which the notion of ``Deep Null Space Learning'' is introduced. More formally, we define data consistency as follows: Given a continuous forward operator $A$ and a noisy measurement $y^\delta := y + \eta = Ax + \eta$, where $\eta$ some noise, could a computed reconstruction $x^\dagger$ have created the measurement $y^\delta$, given the noise level $\|\eta\|$. We call the set of reconstructions that fulfill this property the \textbf{set of valid solutions} and denote it by $V(A, y^\delta, \|\eta\|)$.

In this paper, we present an approach to reconstruction problems that guarantees data consistency by design. We start by reexamining the core challenge of reconstruction problems. We then use the gained insights to propose the projectional method (PM) as a very general framework to tackle inverse problems. A key challenge of the PM lies in the calculation of the projection into the set of valid solutions, $V$. We analyze $V$ and derive a projection algorithm to calculate
\begin{equation}
    P_{V}x^* = \arg\min_{x\in V} \|x-x^*\|,
    \label{eq:set_projection}
\end{equation}
for $A$ being a continuous linear operator. This class of operators includes a wide variety of forward operators like blurring, the Radon transform which plays a prominent role in medical imaging, denoising (the identity), compressed sensing, etc., see e.g., \citet{engl1996regularization, eldar2012compressed}. We also compute the derivative, $\partial P_V$, of $P_V$ via the implicit function theorem, thereby allowing for the usage of $P_V$ as a layer within an end-to-end trained model.

We then, in Section~\ref{sec:red}, demonstrate how the PM not only compares to the recently proposed Regularization by Denoising (RED) (\citet{romano2017little}) but also how RED can be seen as a relaxation of PM. In Section~\ref{sec:dip} we demonstrate how one can apply the projectional approach to Deep Image Prior (DIP) (\citet{ulyanov2018deep}) to avoid the often observed degradation of its reconstructions over time, alleviating the need for early stopping. Finally, in Section~\ref{sec:von_neumann_projection_architecture}, we present a novel neural network architecture based on the PM. The architecture allows for a principled incorporation of prior knowledge from data, while still guaranteeing data consistency -- despite an end-to-end training.
Summary of contributions in order:
\begin{itemize}
    \item A projectional approach to reconstruction that guarantees data consistency.
    \item The derivation of a neural network layer that projects into the set of valid solutions.
    \item Interpretation of RED as an approximation to the projectional method (PM).
    \item Numerical comparison/application of the approach to RED and DIP.
    \item A novel neural network architecture that allows for a principled incorporation of prior knowledge, while still guaranteeing data consistency.
\end{itemize}

\section{A Projectional Ansatz}
\label{sec:a_projectional_ansatz}
\subsection{Motivation and Idea}
We begin with some definitions and notation which we will use throughout the paper:
\begin{itemize}
    \item Let $X$ and $Y$ denote real Hilbert spaces and $A:X\to Y$ a continuous linear operator.
    \item For a given $x\in X$ we define $y := Ax$.
    \item Let further $\delta>0$ be a given \textbf{noise level} and $y^\delta := y + \eta$ be a noisy measurement, where the classical assumption is $\|\eta\| \le \delta$ with $\eta$ some random noise (\citet{engl1996regularization}).
    \item Let $U\subset X$ be a non-empty set, called the \textbf{set of plausible solutions} (e.g.\ sparse or smooth elements or even natural images). For all measurements we will assume $x\in U$.
    \item As discussed above, we informally define the \textbf{set of valid solutions} as the set $V\subset X$ such that the elements of $V(A, y^\delta, \delta)$ would ``explain'' the measurement $y^\delta$, given the operator $A$ and the noise level $\delta$. A formal discussion will follow in Section~\ref{ssec:set_valid_solutions}.
\end{itemize}
Given these definitions we can define the central object of this paper: the set $V\cap U$, which we call the set of \textbf{valid and plausible} solutions. The main goal of this paper is to find an element of $V\cap U$, which we assume to be non-empty.

If we assume $U$ and $V$ to be closed and convex or to have one of many much weaker properties, see \citet{GUBIN19671, Bauschke1993, lewis2008alternating, lewis2009local, drusvyatskiy2015transversality}, etc., we can use \textbf{von Neumann's alternating projection algorithm} (\citet{Bauschke1993}) to accomplish the task of finding an element in $V\cap U$. The algorithm is given by the alternating projections into the two sets $V$ and $U$ and returns in a point in their intersection, i.e.,
\begin{equation}
    U\cap V \ni P_V \circ P_U \circ P_V\circ P_U \circ \cdots \circ P_V x_0,
    \label{eq:alterating}
\end{equation}
for all $x_0\in X$, we always simply set $x_0=0$.
For a visual representation of the algorithm, see Figure~\ref{fig:projection_ball_square} in the appendix.
In this paper, we use this alternating pattern to tackle reconstruction problems. Therefore we rely on having the projections $P_V$ and $P_U$. Since $P_U$ is task-dependent, we begin by analyzing the set of valid solutions and its projection $P_V$.

\subsection{The Set of Valid Solutions}
\label{ssec:set_valid_solutions}
In this subsection, we motivate a definition of the set of valid solutions and analyze it. The classical assumption for inverse problems is that one has $\|\eta\| \le \delta$ for some noise $\eta$ and a noise level $\delta>0$. This could motivate the definition of the set of valid solutions as
\begin{equation}
    \overline V(A, y^\delta, \delta):= \{x\in X:\|Ax-y^\delta\|\le \delta\}.
\end{equation}
But, considering the fact that $\eta$ is usually assumed to be ``unstructured noise'' (e.g. Gaussian) and $X$ is usually assumed to be high- (e.g.\ images) or even infinite-dimensional (e.g.\ functions), we can utilize the principle of concentration of measure (\cite{talagrand1996new}) to conclude that $\|\eta\|\approx\delta$, with increasing accuracy for higher dimensions. This motivates us to define
\begin{equation}
    V(A, y^\delta, \delta):= \{x\in X:\|Ax-y^\delta\| = \delta\}.
\end{equation}
Note that this set is closed since $A$ is continuous. Furthermore, since $A$ is linear, $\overline V$ is also convex. This makes a projection onto $\overline V$ well defined, as well as onto $V$ up to a null set.

In our experiments, however, we found that we rarely had to deal with points in $\overline V$, which makes their projection into $\overline V$ and $V$ equivalent. In practice, we will therefore often simply calculate $P_{\overline V}$ in place of $P_V$, since this can be done very efficiently. We now discuss how to calculate $P_{\overline V}$ and turn it into a neural network layer.

\subsection{Projection into the Valid Solutions as a Layer}
\label{ssec:projection_valid_solution}
In this subsection, we first discuss how to calculate $P_V$ for an $x\notin\overline V$, i.e., $P_{\overline V}$, and then how to calculate its derivative. The calculation of its derivative is of interest since we want to use the projection within a neural network. It would not be feasible to calculate the derivative via some automatic differentiation package, since the calculation of the projection is iterative, which could quickly cause the process to exceed the memory of the machine. From now on, we will assume $x\notin\overline V$.

We begin by deriving an algorithm to calculate the projection into $V,$ i.e., how to solve the constrained optimization problem given by Expression~\eqref{eq:set_projection}.
We can use the method of Lagrangian multipliers to rewrite the expression as the optimization problem
\begin{equation}
    P_V x^* := \arg\min_{x\in X} L_{\mu(\delta, x^*)}(x^*, x),
\end{equation}
where
\begin{equation}
    L_{\mu(\delta, x^*)}(x^*, x) := \frac{\mu(\delta, x^*)}{2} \|Ax-y^\delta\|^2 + \frac{1}{2} \|x-x^*\|^2.
\end{equation}
Since $\mu$ not only depends on $x^*$ but also on $\delta$ via the equality
\begin{equation}
    \|Ax-y^\delta\| = \delta,
\end{equation}
we propose the following alternating algorithm:
\begin{enumerate}
    \setcounter{enumi}{-1}
    \item Find $\mu_+$ s.t. $\varphi(\mu_+)\ge 0$ and $\mu_-$ s.t. $\varphi(\mu_-)\le 0$, set $\mu := \frac{\mu_+ + \mu_-}{2}$.
    \item Solve $x(\mu) := \arg\min_{x\in X} L_{\mu}(x^*, x)$.
    \item Adjust $\mu_+$ and $\mu_-$ via binary search step for root of $\varphi(\mu) := \|Ax(\mu)-y^\delta\|^2 - \delta^2$.
    \item Repeat 1. and 2. until some stopping criterion is reached.
\end{enumerate}
To solve step 1. we can calculate
\begin{equation}
    0 \stackrel{!}{=} F_\mu(x^*, x) := \partial_x L_\mu(x^*, x)^T
    = \left(\mathds{1} + \mu A^TA\right)x - (\mu A^T y^\delta + x^*)
    \label{eq:extrem_point}
\end{equation}
which leads to
\begin{equation}
    x = \left(\mathds{1} + \mu A^TA\right)^{-1}(\mu A^T y^\delta + x^*).
\end{equation}
This can be nicely solved via the efficient conjugate gradient method (CG-method) (\citet{hestenes1952methods}), which does not have to calculate $A^TA$ or even hold it in memory. This can, especially for short-fat matrices (like used in compressed sensing (\citet{eldar2012compressed})), be a significant computational advantage.

The above calculations can be used to flesh out the projection algorithm in more detail, see Algorithm~\ref{algo:ellipsoid_projection}. In practice we stop the computation for $|\|y^\delta - Ax^\dagger\| - \delta| / \delta$ being $\le10^{-2}$, where $x^\dagger$ the current reconstruction. For a complexity analysis of the algorithm see Figure~\ref{fig:complexity_analysis} in the appendix.

\begin{algorithm}[t]
    \SetKwInOut{Input}{Input}
    \SetKwInOut{Output}{Output}

    \underline{function Projection into $\overline V$} $(A,y^\delta, \delta, x^*, \mu_0)$:\\
    \Input{Matrix $A:X\to Y$, vectors $y^\delta \in Y$, $x^* \in X$ and scalars $\delta>0$, $\mu_0>0$}
    \Output{$P_{\overline V} x^*$}
    $\mu_- \leftarrow \mu_0$\\
    $\mu_+ \leftarrow \mu_0$\\

    // Find upper bound for $\mu$.\\
    \While{$\varphi(\mu_-) > 0$} {
        \If {$\mu_- > \mu_+$} {
            $\mu_+ \leftarrow \mu_-$
        }
        $\mu_- \leftarrow 2\mu_-$\\
    }

    // Find lower bound for $\mu$.\\
    \While{$\varphi(\mu_+) < 0$} {
        \If {$\mu_- > \mu_+$} {
            $\mu_- \leftarrow \mu_+$
        }
        $\mu_+ \leftarrow \mu_+ / 2$\\
    }

    // Increase lower bound and decrease upper bound for $\mu$.\\
    $\mu \leftarrow (\mu_- + \mu_+) / 2$\\
    \While{$|\varphi(\mu)| > \epsilon$} {
        \eIf{$\varphi(\mu) \le 0$}{
        $\mu_- \leftarrow \mu$\\
        }{
        $\mu_+ \leftarrow \mu$\\
        }
        $\mu \leftarrow (\mu_- + \mu_+) / 2$
    }
    return $x(\mu)$
    \caption{Projection of a point into the set $\overline V$}
    \label{algo:ellipsoid_projection}
\end{algorithm}

The algorithm allows us, given the projection $P_U$, to solve the inverse problem in the sense of finding a valid and plausible solution via Expression~\eqref{eq:alterating}.

To use $P_{\overline V}$ as a layer in a neural network and to incorporate it in a backpropagation training process, we have to be able to calculate its derivative. As already mentioned, using automatic differentiation may not be feasible, due to the possibly colossal memory requirements caused by the iterative nature of Algorithm~\ref{algo:ellipsoid_projection} (since this effectively would have to be realized via several subsequent layers).

To overcome this problem, we now utilize the implicit function theorem (\citet{krantz2012implicit}) to calculate the derivative of $P_V$ in a way that is also agnostic of the specific algorithm used to calculate the projection itself. This allows the backpropagation procedure to ignore the inner calculations of $P_V$ and treat it essentially as a black box procedure. To be more specific, given $x^*$ and $x := P_V x^*$, we want to calculate $\partial_{x^*}x$.

Since, due to Expression~\eqref{eq:extrem_point}, we have
\begin{align}
    0 &= F_\mu(x^*, x) := \left(\mathds{1} + \mu A^TA\right)x - (\mu A^T y^\delta + x^*)\\
      &= \mu \left(A^TAx - A^T y^\delta\right) + x - x^*,
      \label{eq:fix_point_lambda}
\end{align}
we can calculate the relevant $\mu$ via
\begin{equation}
    \mu = \|x - x^*\| / \|A^T\left(Ax - y^\delta\right)\|,
    \label{eq:mu_equation}
\end{equation}
completely agnostic of the specific algorithm used in the forward propagation to calculate $P_V$.

We can now use the implicit function theorem and, considering Expressions~\eqref{eq:fix_point_lambda} and~\eqref{eq:mu_equation}, obtain
\begin{align}
    \partial_{x^*}x &= -\left(\partial_x F_\mu\right)^{-1} \partial_{x^*} F_\mu\\
                    &= \left(\partial_x F_\mu\right)^{-1}\left(\mathds{1} - \frac{x^* - x}{\|x^* - x\|}\frac{(x^* - x)^T}{\|x^* - x\|}\right)\\
                    &= \left(\mathds{1} + \mu A^TA\right)^{-1}\left(\mathds{1} - \frac{x^* - x}{\|x^* - x\|}\frac{(x^* - x)^T}{\|x^* - x\|}\right)
                    = \left(\partial_{x^*}x\right)^T.
\end{align}
We are now able to calculate $\left(\partial_{x^*}x\right)^Tg^T$, which is necessary of the backpropagation, where $g$ is the gradient flowing downwards from the layer above again -- like in Algorithm~\ref{algo:ellipsoid_projection} -- this also can be done efficiently via the CG-method.

This allows us to implement $P_U$ as a layer in an end-to-end trainable network. We provide code for a PyTorch (\citet{pytorch}) implementation of the projection layer, see \citet{code_projection_layer}.

\subsection{Interpretation}
In this subsection, we discuss interpretations of the PM and its parts and how it relates to variations of the classical $L_2$-regularization
\begin{equation}
    x_{L_2}(\mu) := \arg \min_x \frac{\mu}{2}\|Ax-y^\delta\|^2 + \|Lx\|^2,
    \label{eq:l2_regularization}
\end{equation}
for some regularization parameter $\mu>0$ and some linear continuous operator $L$, often simply $L=id$ (\citet{engl1996regularization}).

We start by interpreting $P_V$ in the $L_2$ context. Usually the regularization parameter $\mu$ is chosen according to the Morozov's Discrepancy Principle (\citet{morozov2012methods}), which, in its classical form states that $\mu$ should be chosen such that (\citet{scherzer1993use})
\begin{equation}
\|Ax_{L_2}(\mu) - y^\delta\| = \delta.
\end{equation}

This leads to the nice interpretation that, for $L=\mathds{1}$, we have
\begin{equation}
    P_V0 = x_{L_2}(\mu(\delta, 0)),
\end{equation}
i.e., $\mu$ is \emph{chosen according to Morozov}.

A further similarity to the $L_2$-regularization becomes apparent when stating the PM via the expression
\begin{equation}
    x_{k+1} := \arg\min_x \frac{\mu(\delta, P_U(x_k))}{2} \|Ax-y^\delta\|^2 + \frac{1}{2} \|x - P_U(x_k)\|^2,
    \label{eq:PM}
\end{equation}
where $x_0 = 0$.
This looks similar to a non-linear version of the non-stationary iterated Tikhonov regularization which is given via
\begin{equation}
    x_{k+1} := \arg\min_x \frac{\mu_k}{2} \|Ax-y^\delta\|^2 + \frac{1}{2} \|Lx - Lx_k\|^2,
    \label{eq:iterated_tikhonov}
\end{equation}
where also $x_0 = 0$ and $\{\mu_k\}_k$ some predetermined sequence (\citet{hanke1998nonstationary}).

\section{Relation to Regularization by Denoising (RED)}
\label{sec:red}
In this section, we want to discuss the connection of the PM with Regularization by Denoising (RED) (\citet{romano2017little}). It could be argued, that the RED (as a direct ``descendent'' of Plug-and-Play priors (\citet{venkatakrishnan2013plug})) and Deep Image Prior (DIP), which we discuss in Section~\ref{sec:dip}, are two of the most relevant recent approaches to regularization.

RED is given by the minimization of the somewhat strange, for technical reasons chosen, functional
\begin{equation}
    L_\text{RED}(\mu) = \frac{\mu}{2}\|Ax-y^\delta\|^2 + x^T\left(x-f(x)\right),
\end{equation}
where $f$ is an arbitrary denoiser such that
\begin{itemize}
    \item $f$ is (locally) positively homogeneous of degree $1$, i.e., $f(cx) = cf(x)$ for $c\ge0$,
    \item $f$ is strongly passive, i.e., the spectral radius of $\partial_x f(x) \le1$ and
    \item $\partial_x f(x)$ is symmetric (\citet{reehorst2018regularization}).
\end{itemize}

An algorithm to solve the minimization problem is given in \citet{romano2017little} and can be expressed via
\begin{equation}
    x_{k+1} := \arg\min_x \frac{\mu}{2} \|Ax-y^\delta\|^2 + \frac{1}{2} \|x - f
    (x_k)\|^2,
    \label{eq:red_iteration}
\end{equation}
for an initial $x_0=0$ and some fixed $\mu>0$.

This means that, if we associate $f$ with $P_U$, RED as stated in Expression~\eqref{eq:red_iteration}, can be seen as a relaxation (setting $\mu$ constant) of the PM, as stated in Expression~\eqref{eq:PM}).

\begin{figure}
    \centering
    \begin{minipage}{.319\textwidth}
    \includegraphics[width=1\linewidth]{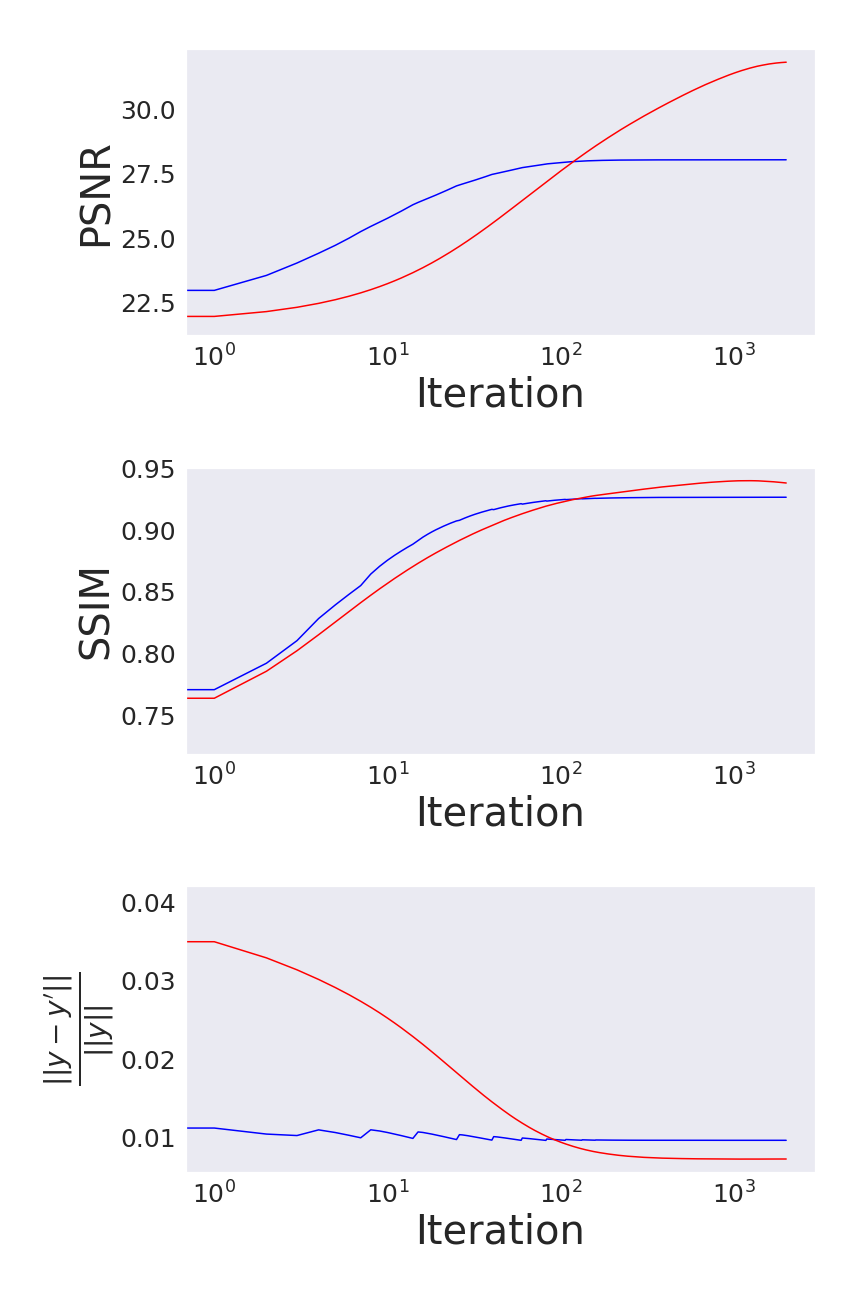}
    \subcaption{1\% noise}
    \end{minipage}
    \begin{minipage}{.319\textwidth}
    \includegraphics[width=1\linewidth]{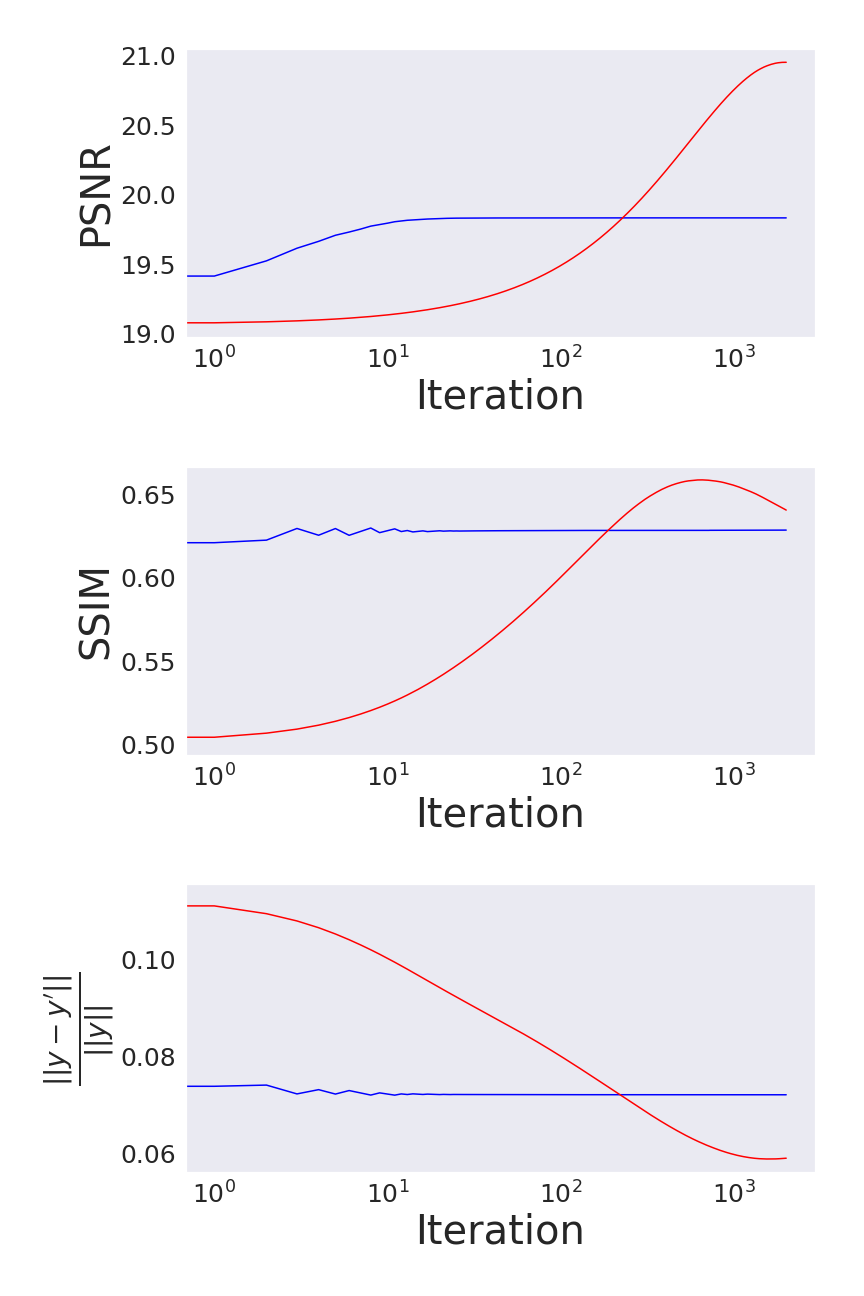}
    \subcaption{10\% noise}
    \end{minipage}
    \begin{minipage}{.319\textwidth}
    \includegraphics[width=1\linewidth]{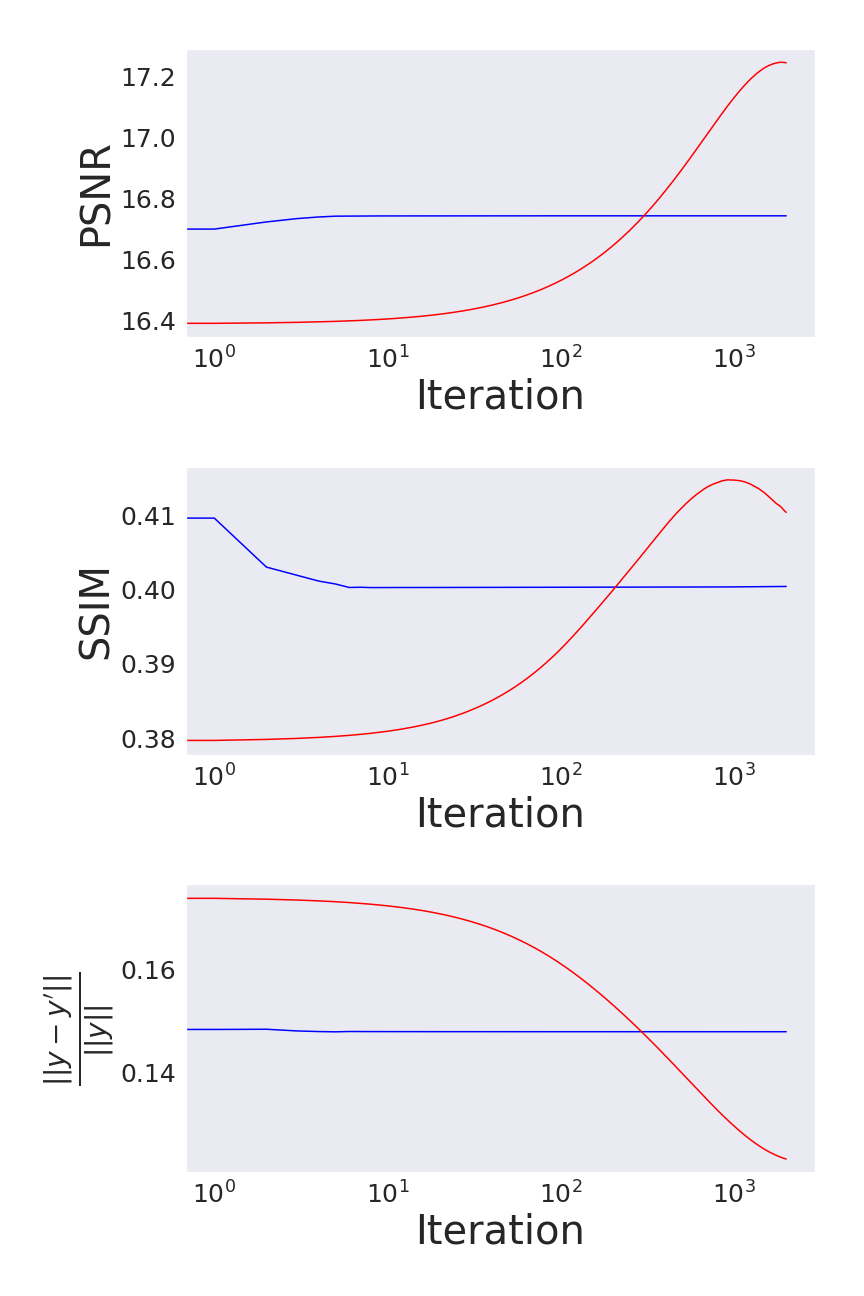}
    \subcaption{30\% noise}
    \end{minipage}
    \caption{The behavior of RED (red) and the PM (blue) over the course of $300$ iterations for reconstructions at three different noise levels. We compare the PSNR, SSIM and $\frac{\|y - y'\|}{\|y\|}$, where $y=Ax$ is the noise free measurement of the ground truth and $y'=Ax^\dagger$ is the noise free measurement of the reconstruction.}
    \label{fig:red_vs_pm}
\end{figure}

\begin{figure}[!htb]
    \centering
    \begin{minipage}{.3\textwidth}
    \includegraphics[trim={41mm 13mm 36mm 13mm}, clip, width=1.\linewidth]{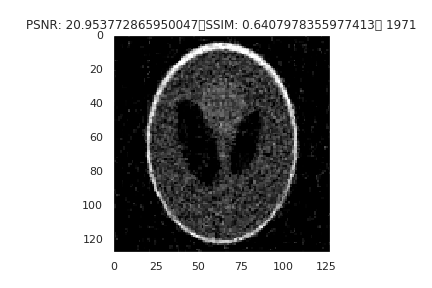}
    \subcaption{RED's best PSNR.\newline PSNR: 20.95, SSIM: 0.64}
    \end{minipage}
    \hfill
    \begin{minipage}{.3\textwidth}
    \includegraphics[trim={41mm 13mm 36mm 13mm}, clip, width=1.\linewidth]{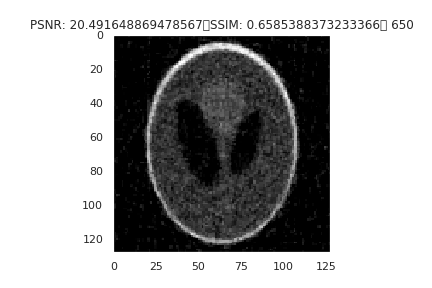}
    \subcaption{RED's best SSIM.\newline PSNR: 20.49, SSIM: 0.66}
    \end{minipage}
    \hfill
    \begin{minipage}{.3\textwidth}
    \vspace{0mm}
    \includegraphics[trim={41mm 13mm 36mm 13mm}, clip, width=1.\linewidth]{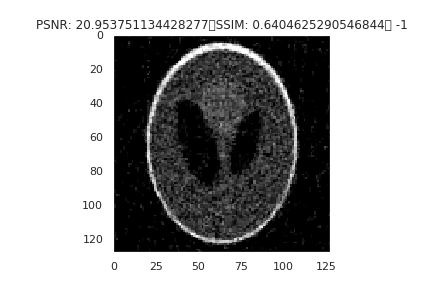}
    \subcaption{RED's last.\newline PSNR: 20.95, SSIM: 0.64}
    \end{minipage}

    \begin{minipage}{.3\textwidth}
    \includegraphics[trim={41mm 13mm 36mm 13mm}, clip, width=1.\linewidth]{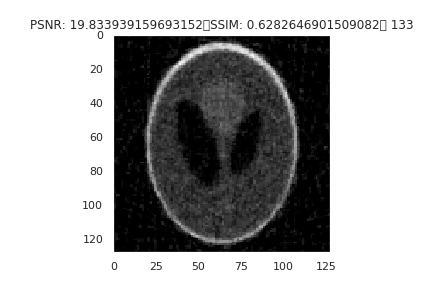}
    \subcaption{PM's best PSNR.\newline PSNR: 19.83 SSIM: 0.63}
    \end{minipage}
    \hfill
    \begin{minipage}{.3\textwidth}
    \includegraphics[trim={41mm 13mm 36mm 13mm}, clip, width=1.\linewidth]{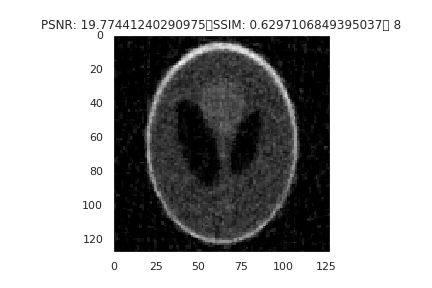}
    \subcaption{PM's best SSIM.\newline PSNR: 19.77, SSIM: 0.63}
    \end{minipage}
    \hfill
    \begin{minipage}{.3\textwidth}
    \vspace{0mm}
    \includegraphics[trim={41mm 13mm 36mm 13mm}, clip, width=1.\linewidth]{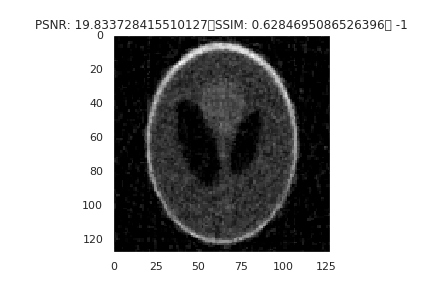}
    \subcaption{PM's last.\newline PSNR: 19.83, SSIM: 0.63}
    \end{minipage}

    \caption{RED and PM reconstructions at $10\%$ noise.}
    \label{fig:red_pm_reconstructions_1e-1}
\end{figure}

\begin{figure}[!htb]
    \centering
    \begin{minipage}{.3\textwidth}
    \includegraphics[trim={41mm 13mm 36mm 13mm}, clip, width=1.\linewidth]{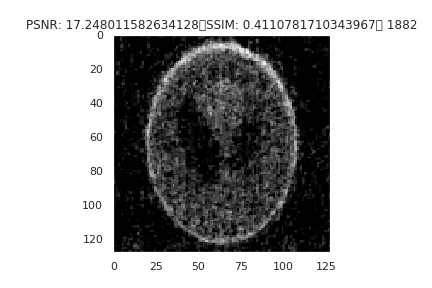}
    \subcaption{RED's best PSNR.\newline PSNR: 17.25, SSIM: 0.41}
    \end{minipage}
    \hfill
    \begin{minipage}{.3\textwidth}
    \includegraphics[trim={41mm 13mm 36mm 13mm}, clip, width=1.\linewidth]{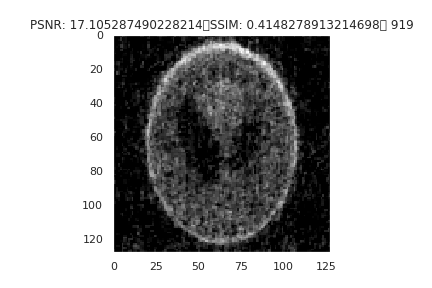}
    \subcaption{RED's best SSIM.\newline PSNR: 17.11, SSIM: 0.41}
    \end{minipage}
    \hfill
    \begin{minipage}{.3\textwidth}
    \vspace{0mm}
    \includegraphics[trim={41mm 13mm 36mm 13mm}, clip, width=1.\linewidth]{figures/reconstructions/red/reconstructions/3e-1_red_last.png}
    \subcaption{RED's last.\newline PSNR: 17.25, SSIM: 0.41}
    \end{minipage}

    \begin{minipage}{.3\textwidth}
    \includegraphics[trim={41mm 13mm 36mm 13mm}, clip, width=1.\linewidth]{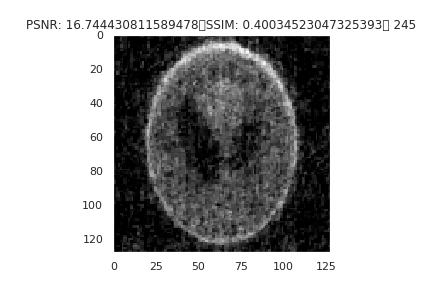}
    \subcaption{PM's best PSNR.\newline PSNR: 16.74 SSIM: 0.4}
    \end{minipage}
    \hfill
    \begin{minipage}{.3\textwidth}
    \includegraphics[trim={41mm 13mm 36mm 13mm}, clip, width=1.\linewidth]{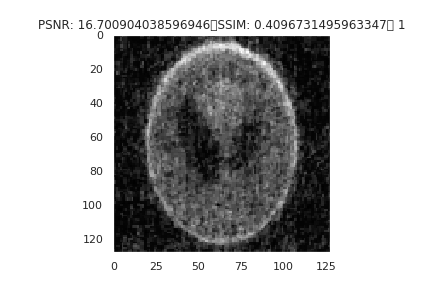}
    \subcaption{PM's best SSIM.\newline PSNR: 17.7 SSIM: 0.4}
    \end{minipage}
    \hfill
    \begin{minipage}{.3\textwidth}
    \vspace{0mm}
    \includegraphics[trim={41mm 13mm 36mm 13mm}, clip, width=1.\linewidth]{figures/reconstructions/red/reconstructions/3e-1_pm_last.png}
    \subcaption{PM's last.\newline PSNR: 17.74 SSIM: 0.4}
    \end{minipage}
    \caption{RED and PM reconstructions at $30\%$ noise.}
    \label{fig:red_pm_reconstructions_3e-1}
\end{figure}

We compared RED and PM over the course of the reconstruction process, see Figure~\ref{fig:red_vs_pm}, where we use, wavelet denoising (\citet{scikit-image, chang2000adaptive}) as $f$ and as an approximation for $P_U$. For our numerical experiments we set $A$ to be the (underdetermined and ill-posed) Radon transform (\citet{helgason1999radon}) with $30$ angles, use the classical Shepp-Logan phantom (\cite{shepp1974fourier}) as the ground truth $x$ and set our noise level $\delta$ to $1\%$, $10\%$ and $30\%$ of the norm of $y = Ax$. We ran each of the experiment $10$ times (different noise), with similar results to the ones shown in Figure~\ref{fig:red_vs_pm}. Unsurprisingly, we see that the RED algorithm performs somewhat better than the PM, since we tuned its hyperparameter over dozens of runs based on the ground truth to maximize the PSNR over $2000$ iterations, whereas the PM was only run once based on the noise level and now tuning with regard to the ground truth in any way.
The $2000$ iterations for RED and for the PM took both on the order of 15 minutes on an Nvidia GeForce GTX 960M. You can find the last reconstructions for a noise level of $30\%$ in Figure~\ref{fig:red_vs_pm_last_rec} and comprehensive comparison of reconstructions in in the appendix, Figure~\ref{fig:red_pm_reconstructions_1e-2}, \ref{fig:red_pm_reconstructions_1e-1} and \ref{fig:red_pm_reconstructions_3e-1}.

\begin{figure}
\centering
\begin{minipage}{.49\textwidth}
    \begin{minipage}{.48\textwidth}
    \includegraphics[trim={41mm 13mm 36mm 13mm}, clip, width=1.\linewidth]{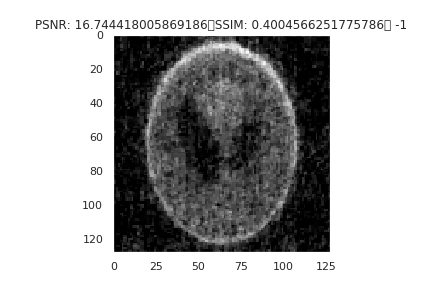}
    \subcaption{PM with a PSNR of 16.74 and a SSIM of 0.4.}
    \end{minipage}
    \hfill
    \begin{minipage}{.48\textwidth}
    \includegraphics[trim={41mm 13mm 36mm 13mm}, clip, width=1.\linewidth]{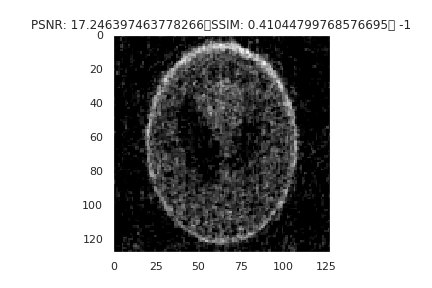}
    \subcaption{RED with a PSNR of 17.25 and a SSIM of 0.41.}
    \end{minipage}
    \caption{The last PM and RED reconstructions for a noise level of $30\%$}
    \label{fig:red_vs_pm_last_rec}
\end{minipage}
\hfill
\begin{minipage}{.49\textwidth}
    \begin{minipage}{.48\textwidth}
    \includegraphics[trim={14mm 10mm 23mm 16mm}, clip, width=1.\linewidth]{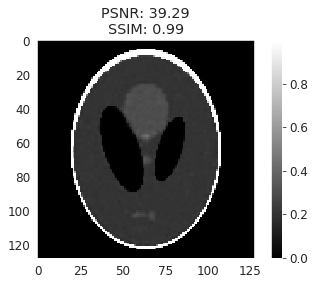}
    \subcaption{DIP+$\delta$ with a PSNR of 39.29 and a SSIM of 0.99.}
    \end{minipage}
    \hfill
    \begin{minipage}{.48\textwidth}
    \includegraphics[trim={14mm 10mm 23mm 16mm}, clip, width=1.\linewidth]{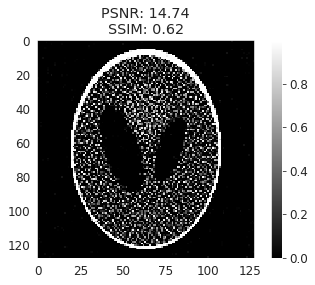}
    \subcaption{DIP with a PSNR of 14.74 and a SSIM of 0.62.}
    \end{minipage}
    \caption{The last DIP+$\delta$ and DIP reconstructions for a noise level of $1\%$}
    \label{fig:dip_dipr_last_rec}
\end{minipage}
\end{figure}

\section{Application to Deep Image Prior (DIP)}
\label{sec:dip}
In this section, we discuss how one can implement the central idea of finding an element in $U\cap V$ for the case where $U$ is given by all possible reconstructions from a Deep Image Prior (DIP).

The idea of using DIP to solve inverse problems, as described by \citet{van2018compressed}, is to minimize the functional
\begin{equation}
    L_\text{DIP}(\Theta) := \frac{1}{2}\|AG_\Theta(z) - y^\delta\|^2
\end{equation}
with regard to $\Theta$, where $G_\Theta(z)$ is an untrained neural network with a fixed random input $z$ and parameters $\Theta$. The reconsturction is then given via $G_\Theta(z)$, i.e., the output of the network.

We propose the following simple modification to the DIP functional based on the idea to find a valid and plausible reconstruction. Specifically, we propose
\begin{equation}
    L_\text{DIP+$\delta$}(\Theta) := \left(\|AG_\Theta(z) - y^\delta\|^2 - \delta^2\right)^2.
\end{equation}
Here the reconstruction is given by $G_\Theta(z)$ after the minimization.

\begin{figure}
    \centering
    \begin{minipage}{.319\textwidth}
    \includegraphics[width=1\linewidth]{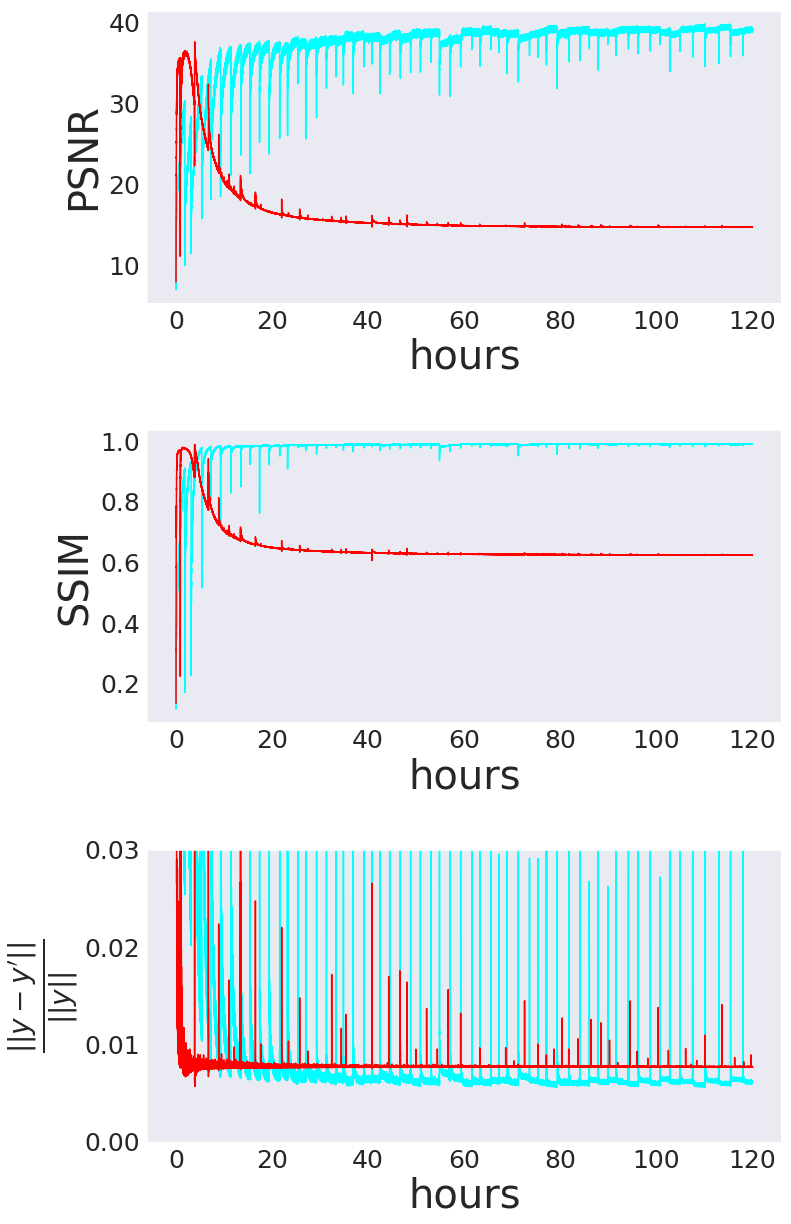}
    \subcaption{1\% noise}
    \end{minipage}
    \begin{minipage}{.319\textwidth}
    \includegraphics[width=1\linewidth]{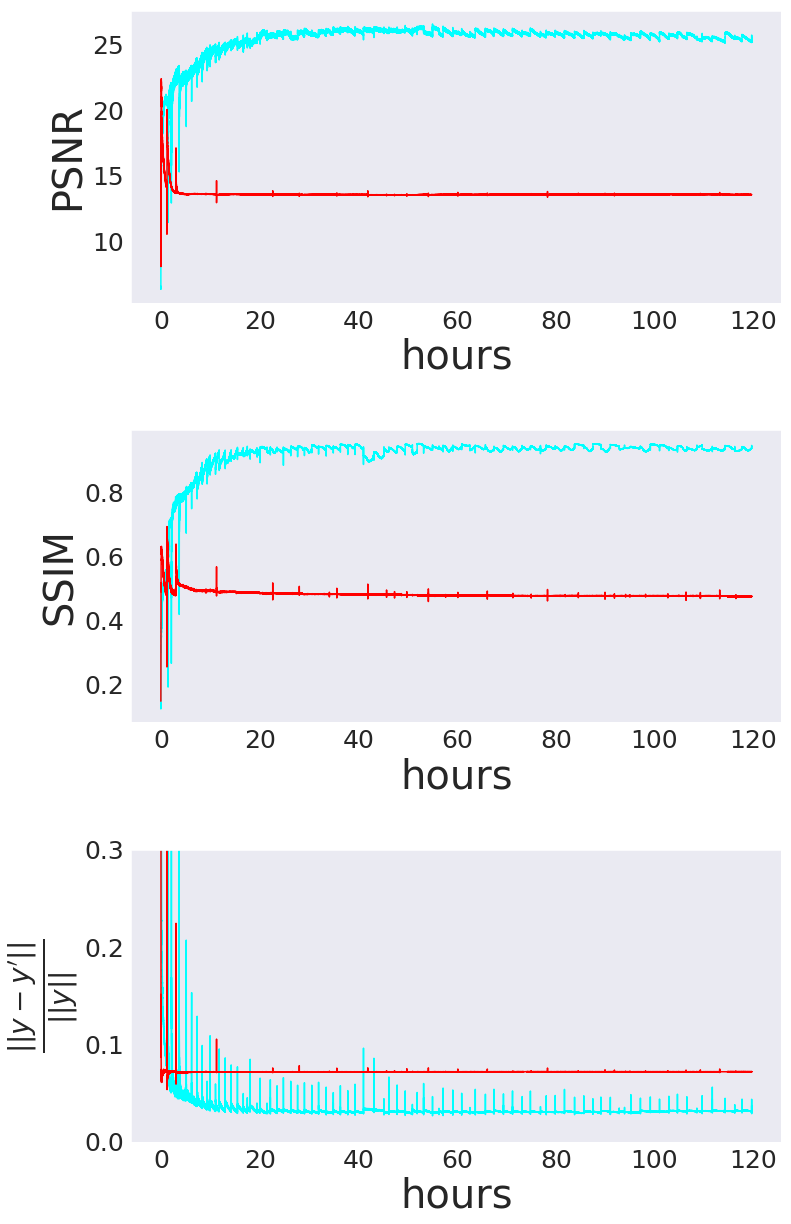}
    \subcaption{10\% noise}
    \end{minipage}
    \begin{minipage}{.319\textwidth}
    \includegraphics[width=1\linewidth]{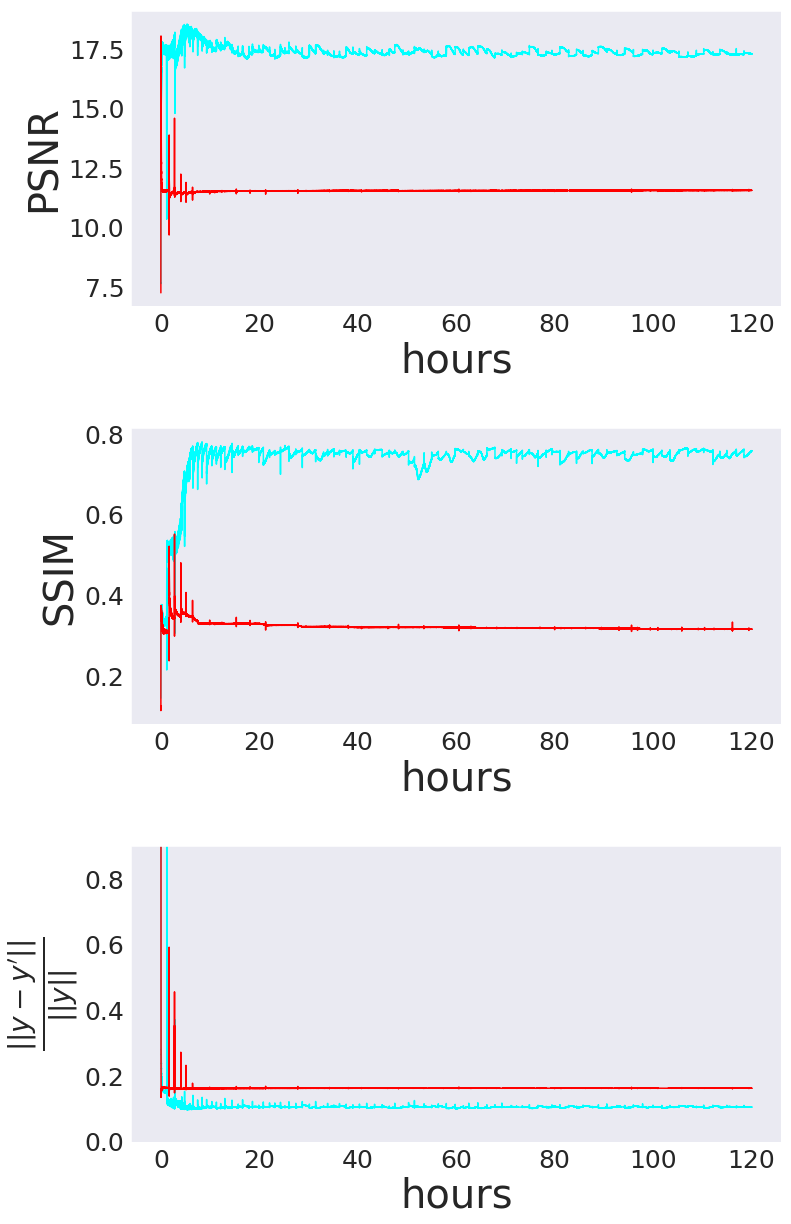}
    \subcaption{30\% noise}
    \end{minipage}
    \caption{The behavior of the minimization of $L_\text{DIP}$ (red) and of $L_\text{DIP+$\delta$}$ (cyan) over the course of the reconstruction.}
    \label{fig:dip_results}
\end{figure}

\begin{figure}[!htb]
    \centering
    \begin{minipage}{.3\textwidth}
    \includegraphics[trim={11mm 8mm 20mm 15mm}, clip, width=1.\linewidth]{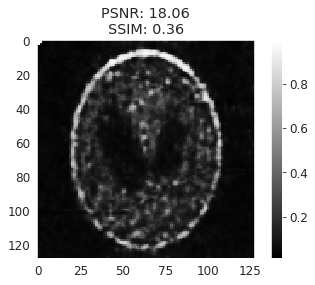}
    \subcaption{DIP's best PSNR.\newline PSNR: 18.06, SSIM: 0.36}
    \end{minipage}
    \hfill
    \begin{minipage}{.3\textwidth}
    \includegraphics[trim={11mm 8mm 20mm 15mm}, clip, width=1.\linewidth]{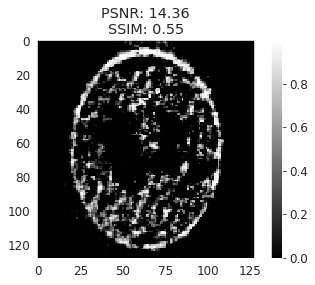}
    \subcaption{DIP's best SSIM.\newline PSNR: 14.36, SSIM: 0.55}
    \end{minipage}
    \hfill
    \begin{minipage}{.3\textwidth}
    \vspace{0mm}
    \includegraphics[trim={11mm 8mm 20mm 15mm}, clip, width=1.\linewidth]{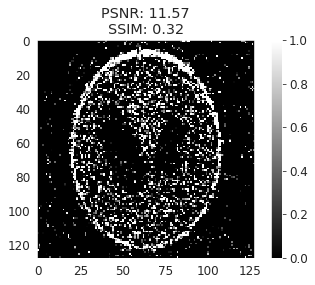}
    \subcaption{DIP's last.\newline PSNR: 11.57, SSIM: 0.32}
    \end{minipage}

    \begin{minipage}{.3\textwidth}
    \includegraphics[trim={11mm 8mm 20mm 15mm}, clip, width=1.\linewidth]{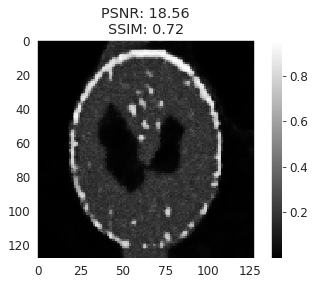}
    \subcaption{DIP+$\delta$'s best PSNR.\newline PSNR: 18.56, SSIM: 0.72}
    \end{minipage}
    \hfill
    \begin{minipage}{.3\textwidth}
    \includegraphics[trim={11mm 8mm 20mm 15mm}, clip, width=1.\linewidth]{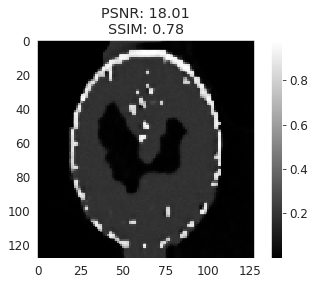}
    \subcaption{DIP+$\delta$'s best SSIM.\newline PSNR: 18.01, SSIM: 0.78}
    \end{minipage}
    \hfill
    \begin{minipage}{.3\textwidth}
    \vspace{0mm}
    \includegraphics[trim={11mm 8mm 20mm 15mm}, clip, width=1.\linewidth]{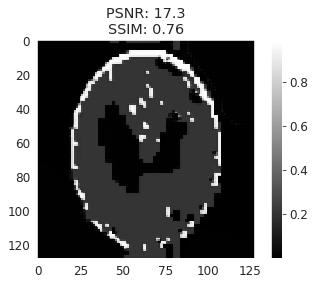}
    \subcaption{DIP+$\delta$'s last.\newline PSNR: 17.3, SSIM: 0.76}
    \end{minipage}

    \caption{DIP and DIP+$\delta$ reconstructions at $30\%$ noise.}
    \label{fig:dip_reconstructions_3e-1}
\end{figure}

We now numerically compare this modified functional $L_\text{DIP+$\delta$}$ with $L_\text{DIP}$ in the same setup as used in Section~\ref{sec:red} for RED and the PM based on wavelet denoising. As the DIP network, we use the ``skip net'' UNet described in the original DIP paper (\citet{ulyanov2018deep}) and Adam (\citet{kingma2014adam}) with standard settings (of PyTorch), which we found to work best for the DIP reconstructions. The results can be found in Figure~\ref{fig:dip_results}. Like in Section~\ref{sec:red} we compare the PSNR, SSIM and $\|y - y'\| / \|y\|$, where $y=Ax$ is the noise-free measurement of the ground truth and $y'=Ax^\dagger$ the noise-free measurement of the reconstruction. We plot the reconstruction over time, not over the number of iterations, to clearly show that even over relatively long time scales the solutions of $L_\text{DIP+$\delta$}$ do not exhibit signs of much deterioration. We ran the experiments in parallel on two Nvidia GeForce GTX 1080 ti.
You can find the last reconstructions for a noise level of $1\%$ in Figure~\ref{fig:dip_dipr_last_rec} and comprehensive comparison of reconstructions in in the appendix, Figure~\ref{fig:dip_reconstructions_1e-2}, \ref{fig:dip_reconstructions_1e-1} and \ref{fig:dip_reconstructions_3e-1}. All experiments reached approximately $750,000$ iterations over the course of the $120$ hours of the experiment. We find that the modified functional shows much less deterioration of the reconstruction overtime and outperforms the vanilla DIP (except for short spikes) in all our metrics.

\section{von Neumann Projection Architecture}
\label{sec:von_neumann_projection_architecture}
In this section, we discuss how one can learn, in an end-to-end fashion, a replacement for $P_U$, where $U$ an arbitrary set. For this, we heavily rely on the fact that we can calculate the gradients of $P_U$ and can, therefore, use it as a layer (see Section~\ref{ssec:projection_valid_solution}).

Based on Expression~\eqref{eq:alterating} we propose the following neural network architecture:
\begin{equation}
    G_\theta(A, y^\delta, \delta, x_0) := P_{\overline V(A, y^\delta, \delta)} \circ g_{\theta, n-1} \circ P_{\overline V(A, y^\delta, \delta)} \circ \cdots \circ g_{\theta, 0} \circ P_{\overline V(A, y^\delta, \delta)} (x_0),
    \label{eq:vNPA}
\end{equation}
where the $g_{\theta, i}: X \to X$ are neural networks, e.g.\ UNets (\citet{ronneberger2015u}) or autoencoders (\citet{ng2011sparse}), for simplicity we set $x_0=0$.

This type of alternating architecture, which we call \textbf{von Neumann projection architecture} (vNPA), allows one to use the PM while incorporating prior information from data in a highly customized manner. This is in stark contrast to the use of usually quite general Plug-and-Play priors (\citet{venkatakrishnan2013plug, sreehari2016plug}), which are not specifically adapted to the reconstruction task. Unlike other end-to-end approaches this architecture, using $P_V$, guarantees the data consistency of the reconstruction.

We now demonstrate the approach on a toy example. For that we set $U$ to be a set of low frequency functions $f:[0,127] \ni x \mapsto f(x) \in [0,1]$, which for some random $x\ge63$ turn continuously into linear functions such that $f(127) = 1$, see Figure~\ref{fig:test_function} in the appendix for example plots and Python code that creates random instances of these functions in the form of vectors of the length $128$. One could try to design a handcrafted prior for these arbitrarily chosen functions, but it would be easier to simply learn one. As the forward operator, $A\in\mathbb{R}^{64\times128}$, we use a typical compressed sensing operator, i.e., the entries are samples from a standard Gaussian distribution.

We use the vNPA as described in Expression~\eqref{eq:vNPA}, with $n=4$. Each $g_{\theta, i}$ is a separate vanilla autoencoder with the widths $128\rightarrow64\rightarrow32\rightarrow64\rightarrow128$, where all layers up to the last one are vanilla ReLU layer, the last one being a vanilla Sigmoid layer. We trained it with a batch size of $32$ and Adam (\citet{kingma2014adam}) at a learning rate of $10^{-2}$ for $2,000$ batches over 45 minutes on an Nvidia GeForce GTX 960M at a noise level of $1\%$.

\begin{figure}
\centering
\begin{minipage}{.24\textwidth}
    \centering
    \includegraphics[trim={0cm 0cm 38cm 7.5mm},clip, width=1.\linewidth]{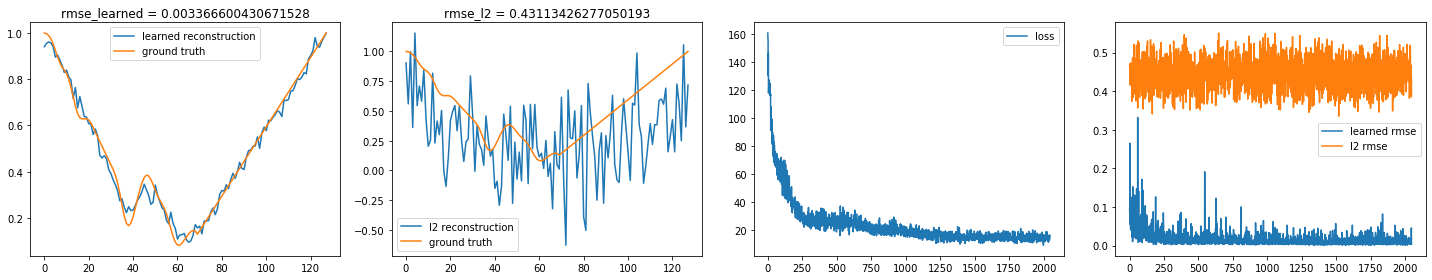}
\end{minipage}
\hfill
\begin{minipage}{.24\textwidth}
    \centering
    \includegraphics[trim={0cm 0cm 38cm 7.5mm},clip, width=1.\linewidth]{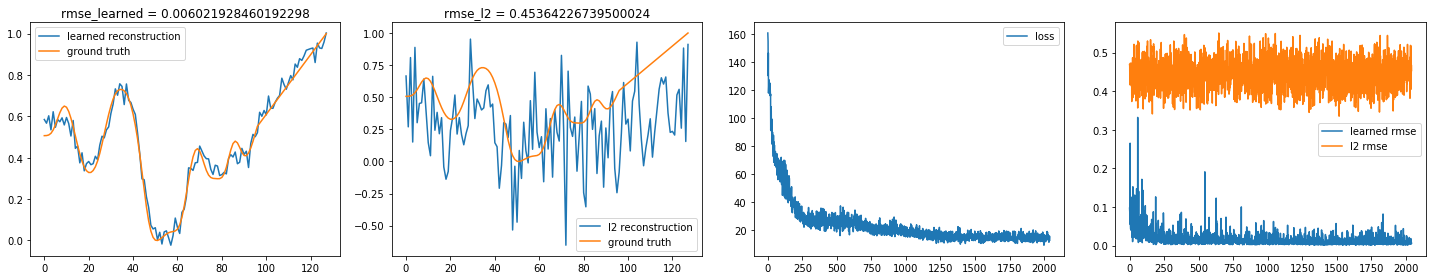}
\end{minipage}
\hfill
\begin{minipage}{.24\textwidth}
    \centering
    \includegraphics[trim={0cm 0cm 38cm 7.5mm},clip, width=1.\linewidth]{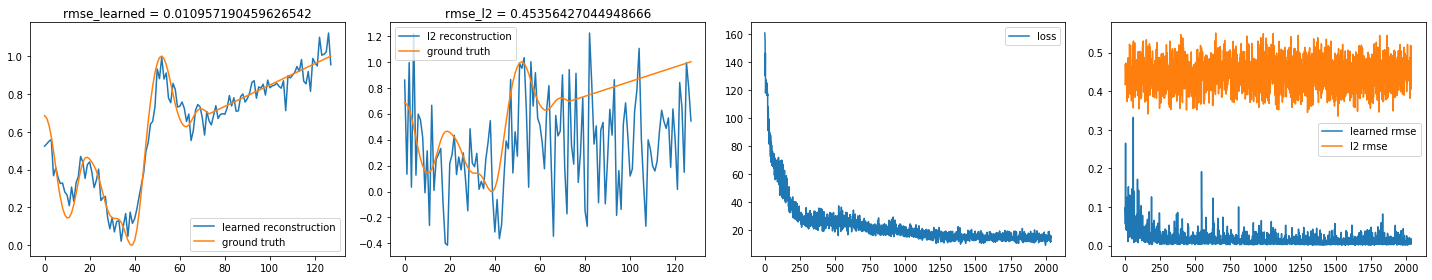}
\end{minipage}
\hfill
\begin{minipage}{.24\textwidth}
    \centering
    \includegraphics[trim={0cm 0cm 38cm 7.5mm},clip, width=1.\linewidth]{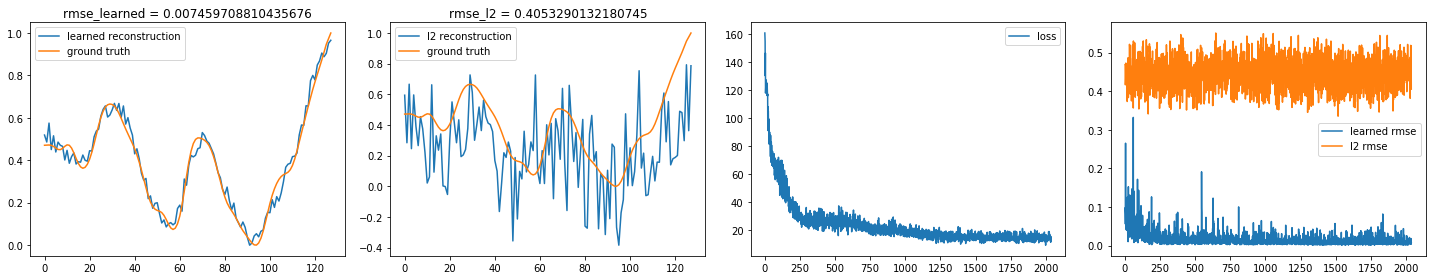}
\end{minipage}

\begin{minipage}{.24\textwidth}
    \centering
    \includegraphics[trim={12.6cm 0cm 25.4cm 7.5mm},clip, width=1.\linewidth]{figures/reconstructions/test_functions/rec-5.png}
    \subcaption{Errors: $0.0034$ vs $0.43$}
\end{minipage}
\hfill
\begin{minipage}{.24\textwidth}
    \centering
    \includegraphics[trim={12.6cm 0cm 25.4cm 7.5mm},clip, width=1.\linewidth]{figures/reconstructions/test_functions/rec-9.png}
    \subcaption{Errors: $0.006$ vs $0.45$}
\end{minipage}
\hfill
\begin{minipage}{.24\textwidth}
    \centering
    \includegraphics[trim={12.6cm 0cm 25.4cm 7.5mm},clip, width=1.\linewidth]{figures/reconstructions/test_functions/rec-11.png}
    \subcaption{Errors: $0.011$ vs $0.45$}
\end{minipage}
\hfill
\begin{minipage}{.24\textwidth}
    \centering
    \includegraphics[trim={12.6cm 0cm 25.4cm 7.5mm},clip, width=1.\linewidth]{figures/reconstructions/test_functions/rec-12.png}
    \subcaption{Errors: $0.007$ vs $0.41$}
\end{minipage}
\caption{Learned (top) v.s.\ best mean-squared-error $l_2$-reconstructions (bottom) with there respective relative $l_2$-errors, i.e., $\frac{\|x-x_\text{reconstruction}\|}{\|x\|}$.}
\label{fig:test_function_reconstructions}
\end{figure}

Example reconstructions via the network compared to $L_2$ reconstructions (the regularization parameter optimized such that it minimizes the $L_2$ error to the ground truth) can be found in Figure~\ref{fig:test_function_reconstructions}. Each took less than half a second to compute. The network produces better reconstruction than the optimized $L_2$ regularization.

\FloatBarrier

\section{Conclusion}
We proposed a projectional approach of optimizing for an element that is valid and plausible, given the operator, measurement, and noise level. We find that the approach is fruitful and widely applicable. Specifically, we demonstrated its applicability on the one hand via very general Plug-and-Play priors like wavelet denoising or DIP and on the other hand via highly task-specific learned priors via the von Neumann projection architecture.

We also show how the approach can be connected to the well studied iterated Tikhonov reconstructions, how it allows for an interpretation of the somewhat strange, but highly effective RED functional and how it can be used to stabilize and improve DIP reconstructions.


\bibliographystyle{plainnat}
\bibliography{neurips_2019}

\clearpage
\section*{Appendix}
\FloatBarrier
\begin{figure}[h]
\centering
    \centering
    \includegraphics[width=0.7\linewidth]{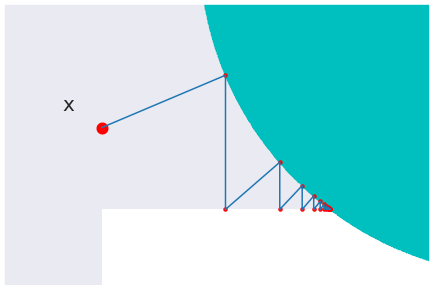}
    \caption{A visualization of von Neumann's projection algorithm: mapping the point $x$ onto the intersection of a disk (cyan) and a square (white).}
    \label{fig:projection_ball_square}
\end{figure}

\begin{figure}[h]
    \centering
    \begin{minipage}{.44\textwidth}
    \includegraphics[width=1.\linewidth]{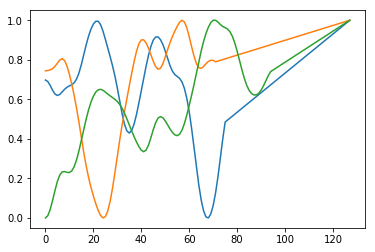}
    \end{minipage}
    \begin{minipage}{.53\textwidth}
    \vspace{-1mm}
\begin{lstlisting}[language=Python]
import numpy as np

def get_function_sample(function_length=128):
    # Create random frequency amplitudes
    freq = np.random.normal(0, 1, function_length + 1)
    # Dampen higher frequencies
    freq *= np.logspace(0, -8, function_length + 1)
    # apply inverse real Fourier trainsform
    f = np.fft.irfft(freq)
    # Remove symmetric part
    f = f[:function_length]
    # Normalize between 0 and 1
    f -= np.min(f)
    f /= np.max(f)

    # Generate random point to begin linear part 
    start_lin = int(np.random.rand() * 64 + 64)
    # Generate linear part
    f[start_lin:] = np.linspace(f[start_lin], 1, len(f[start_lin:]))

    return f
\end{lstlisting}
\end{minipage}
\caption{Three example test functions and the Python code that creates one.}
\label{fig:test_function}
\end{figure}

\begin{figure}[h]
    The complexity of Algorithm~\ref{algo:ellipsoid_projection}, can simply be estimated via the complexity of the binary search method, $O\left(\left(\log{\frac{1}{\epsilon}}\right)F(\epsilon)\right)$, where $\epsilon>0$ is the required precision and $F$ is the complexity of the CG-method applied to $M$. Assuming $\kappa$ is the condition number of $M$ the complexity of the CG-method is $O(\sqrt{\kappa}\log{\frac{1}{\epsilon})}$ (\citet{saad2003iterative}). This results in an overall complexity of
\begin{equation}
    O\left(\sqrt{\kappa}\left(\log{\frac{1}{\epsilon}}\right)^2\right).
\end{equation}
    \caption{A complexity analysis of Algorithm~\ref{algo:ellipsoid_projection}.}
    \label{fig:complexity_analysis}
\end{figure}


\begin{figure}[!htb]
    \centering
    \begin{minipage}{.3\textwidth}
    \includegraphics[trim={41mm 13mm 36mm 13mm}, clip, width=1.\linewidth]{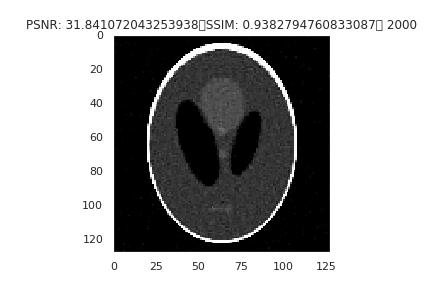}
    \subcaption{RED's best PSNR.\newline PSNR: 31.84, SSIM: 0.94}
    \end{minipage}
    \hfill
    \begin{minipage}{.3\textwidth}
    \includegraphics[trim={41mm 13mm 36mm 13mm}, clip, width=1.\linewidth]{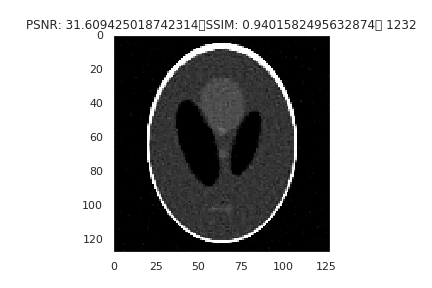}
    \subcaption{RED's best SSIM.\newline PSNR: 31.61, SSIM: 0.94}
    \end{minipage}
    \hfill
    \begin{minipage}{.3\textwidth}
    \vspace{0mm}
    \includegraphics[trim={41mm 13mm 36mm 13mm}, clip, width=1.\linewidth]{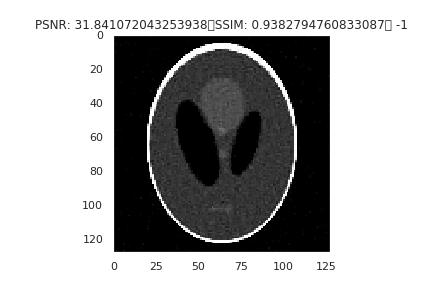}
    \subcaption{RED's last.\newline PSNR: 31.84, SSIM: 0.94}
    \end{minipage}

    \begin{minipage}{.3\textwidth}
    \includegraphics[trim={41mm 13mm 36mm 13mm}, clip, width=1.\linewidth]{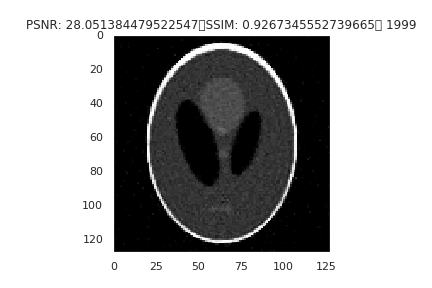}
    \subcaption{PM's best PSNR.\newline PSNR: 28.05 SSIM: 0.93}
    \end{minipage}
    \hfill
    \begin{minipage}{.3\textwidth}
    \includegraphics[trim={41mm 13mm 36mm 13mm}, clip, width=1.\linewidth]{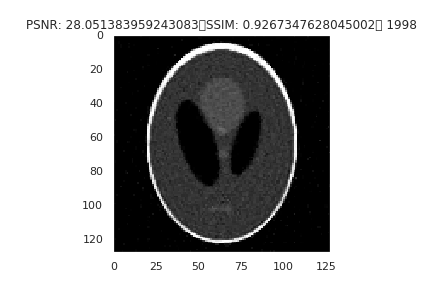}
    \subcaption{PM's best SSIM.\newline PSNR: 28.05, SSIM: 0.93}
    \end{minipage}
    \hfill
    \begin{minipage}{.3\textwidth}
    \vspace{0mm}
    \includegraphics[trim={41mm 13mm 36mm 13mm}, clip, width=1.\linewidth]{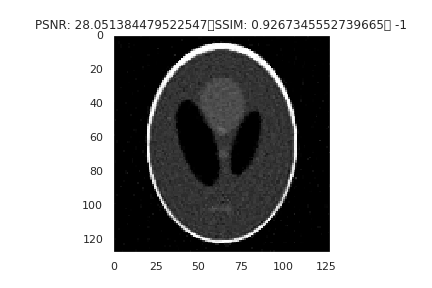}
    \subcaption{PM's last.\newline PSNR: 28.05, SSIM: 0.93}
    \end{minipage}

    \caption{RED and PM reconstructions at $1\%$ noise.}
    \label{fig:red_pm_reconstructions_1e-2}
\end{figure}

\begin{figure}[!htb]
    \centering
    \begin{minipage}{.3\textwidth}
    \includegraphics[trim={11mm 8mm 20mm 15mm}, clip, width=1.\linewidth]{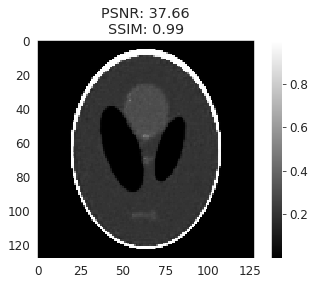}
    \subcaption{DIP's best PSNR.\newline PSNR: 37.66, SSIM: 0.99}
    \end{minipage}
    \hfill
    \begin{minipage}{.3\textwidth}
    \includegraphics[trim={11mm 8mm 20mm 15mm}, clip, width=1.\linewidth]{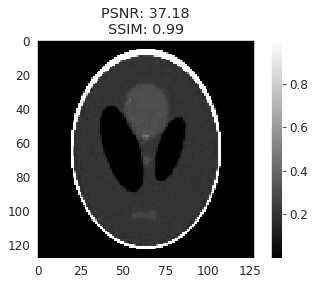}
    \subcaption{DIP's best SSIM.\newline PSNR: 37.18, SSIM: 0.99}
    \end{minipage}
    \hfill
    \begin{minipage}{.3\textwidth}
    \vspace{0mm}
    \includegraphics[trim={11mm 8mm 20mm 15mm}, clip, width=1.\linewidth]{figures/reconstructions/dip/reconstructions/dip_last_01.png}
    \subcaption{DIP's last.\newline PSNR: 14.74, SSIM: 0.62}
    \end{minipage}

    \begin{minipage}{.3\textwidth}
    \includegraphics[trim={11mm 8mm 20mm 15mm}, clip, width=1.\linewidth]{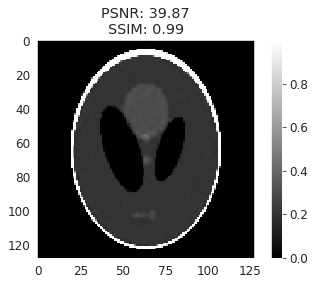}
    \subcaption{DIP+$\delta$'s best PSNR.\newline PSNR: 39.87, SSIM: 0.99}
    \end{minipage}
    \hfill
    \begin{minipage}{.3\textwidth}
    \includegraphics[trim={11mm 8mm 20mm 15mm}, clip, width=1.\linewidth]{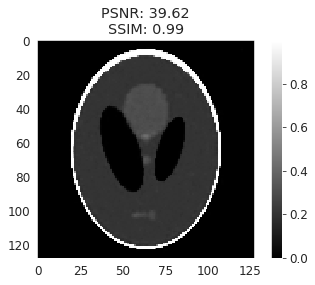}
    \subcaption{DIP+$\delta$'s best SSIM.\newline PSNR: 39.62, SSIM: 0.99}
    \end{minipage}
    \hfill
    \begin{minipage}{.3\textwidth}
    \vspace{0mm}
    \includegraphics[trim={11mm 8mm 20mm 15mm}, clip, width=1.\linewidth]{figures/reconstructions/dip/reconstructions/dipr_last_01.png}
    \subcaption{DIP+$\delta$'s last.\newline PSNR: 39.29, SSIM: 0.99}
    \end{minipage}

    \caption{DIP and DIP+$\delta$ reconstructions at $1\%$ noise.}
    \label{fig:dip_reconstructions_1e-2}
\end{figure}

\begin{figure}[!htb]
    \centering
    \begin{minipage}{.3\textwidth}
    \includegraphics[trim={11mm 8mm 20mm 15mm}, clip, width=1.\linewidth]{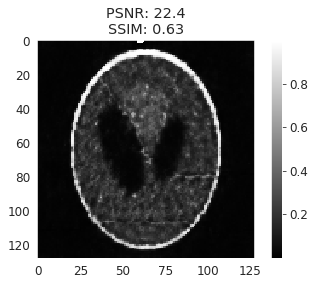}
    \subcaption{DIP's best PSNR.\newline PSNR: 22.4, SSIM: 0.63}
    \end{minipage}
    \hfill
    \begin{minipage}{.3\textwidth}
    \includegraphics[trim={11mm 8mm 20mm 15mm}, clip, width=1.\linewidth]{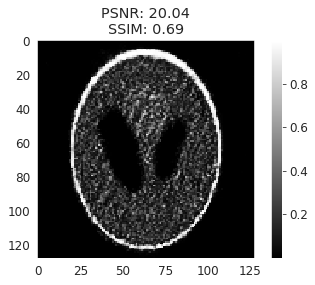}
    \subcaption{DIP's best SSIM.\newline PSNR: 20.04, SSIM: 0.69}
    \end{minipage}
    \hfill
    \begin{minipage}{.3\textwidth}
    \vspace{0mm}
    \includegraphics[trim={11mm 8mm 20mm 15mm}, clip, width=1.\linewidth]{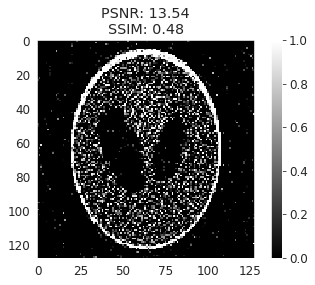}
    \subcaption{DIP's last.\newline PSNR: 13.54, SSIM: 0.48}
    \end{minipage}

    \begin{minipage}{.3\textwidth}
    \includegraphics[trim={11mm 8mm 20mm 15mm}, clip, width=1.\linewidth]{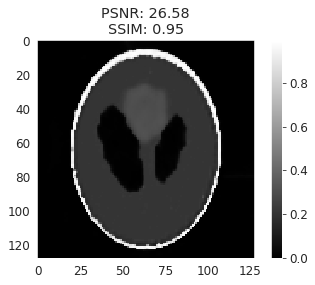}
    \subcaption{DIP+$\delta$'s best PSNR.\newline PSNR: 26.58, SSIM: 0.95}
    \end{minipage}
    \hfill
    \begin{minipage}{.3\textwidth}
    \includegraphics[trim={11mm 8mm 20mm 15mm}, clip, width=1.\linewidth]{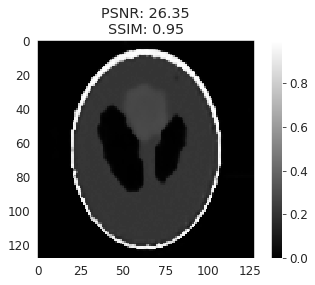}
    \subcaption{DIP+$\delta$'s best SSIM.\newline PSNR: 26.35, SSIM: 0.95}
    \end{minipage}
    \hfill
    \begin{minipage}{.3\textwidth}
    \vspace{0mm}
    \includegraphics[trim={11mm 8mm 20mm 15mm}, clip, width=1.\linewidth]{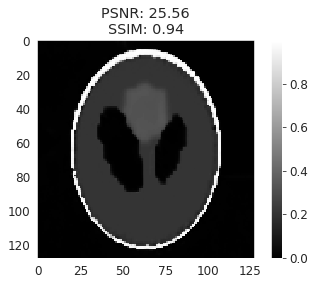}
    \subcaption{DIP+$\delta$'s last.\newline PSNR: 25.56, SSIM: 0.94}
    \end{minipage}

    \caption{DIP and DIP+$\delta$ reconstructions at $10\%$ noise.}
    \label{fig:dip_reconstructions_1e-1}
\end{figure}

\end{document}